\documentclass[conference]{IEEEtran}
\IEEEoverridecommandlockouts
\usepackage{cite}
\usepackage{amsmath,amssymb,amsfonts}
\usepackage{algorithmic}
\usepackage{graphicx}
\usepackage{textcomp}
\usepackage{xcolor}
\usepackage{mathtools}
\usepackage{bm}
\usepackage{makecell}
\usepackage{multirow}
\usepackage{booktabs}
\usepackage[flushleft]{threeparttable}
\usepackage{orcidlink}
\usepackage[switch]{lineno}
\usepackage{xspace}
\usepackage[caption=false]{subfig}
\usepackage[inline]{enumitem}

\newcommand{\name}{ViRED\xspace}
\def\BibTeX{{\rm B\kern-.05em{\sc i\kern-.025em b}\kern-.08em
    T\kern-.1667em\lower.7ex\hbox{E}\kern-.125emX}}
\begin{document}

\title{ViRED: Prediction of Visual Relations in Engineering Drawings\\
}

\author{

\IEEEauthorblockN{
Chao Gu \orcidlink{0009-0004-6321-0475}\IEEEauthorrefmark{1},
Ke Lin \orcidlink{0009-0002-5376-7881}\IEEEauthorrefmark{2},
Yiyang Luo \orcidlink{0009-0008-8094-180X}\IEEEauthorrefmark{3},
Jiahui Hou \orcidlink{0000-0002-3340-8585}\IEEEauthorrefmark{1},
and Xiang-Yang Li \orcidlink{0000-0002-6070-6625}\IEEEauthorrefmark{1}
}





	\IEEEauthorblockA{\IEEEauthorrefmark{1}School of Computer Science and Technology and LINKE Lab, University of Science and Technology of China, Hefei, China}

	\IEEEauthorblockA{\IEEEauthorrefmark{2}School of Software, Tsinghua University, Beijing, China}

    \IEEEauthorblockA{\IEEEauthorrefmark{3}School of Computer Science and Engineering, Nanyang Technological University, Singapore, Singapore}

    \IEEEauthorblockA{guch8017@mail.ustc.edu.cn,\{leonard.keilin,lawrence.luoyy\}@gmail.com,\{jhhou,xiangyangli\}@ustc.edu.cn}

}


\maketitle

\begin{abstract}
To accurately understand engineering drawings, it is essential to establish the correspondence between images and their description tables within the drawings. 
Existing document understanding methods predominantly focus on text as the main modality, which is not suitable for documents containing substantial image information. 
In the field of visual relation detection, the structure of the task inherently limits its capacity to assess relationships among all entity pairs in the drawings. 

To address this issue, we propose a vision-based relation detection model, named \name, to identify the associations between tables and circuits in electrical engineering drawings. 
Our model mainly consists of three parts: a vision encoder, an object encoder, and a relation decoder. 

We implement \name using PyTorch to evaluate its performance. To validate the efficacy of \name, we conduct a series of experiments. 
The experimental results indicate that, within the engineering drawing dataset, our approach attained an accuracy of 96\% in the task of relation prediction, marking a substantial improvement over existing methodologies. The results also show that \name can inference at a fast speed even when there are numerous objects in a single engineering drawing.
\end{abstract}


\section{Introduction}

\IEEEPARstart{D}{igitization} of engineering design drawings constitutes a crucial component of contemporary industrial processes.
Nevertheless, the automated recognition of these engineering drawings in image format still encounters considerable challenges. 
Electrical engineering drawings, a subset of engineering drawings, are primarily used to depict equipment related to electrical systems. 
Typically, electrical design engineers are required to review and recreate numerous electrical engineering drawings to transform technical illustrations into production-level drawings. 
To alleviate this workload, it is crucial to develop automated recognition methods for electrical engineering drawings, which are stored in image format.

\begin{figure*}[t]
    \centering
    \includegraphics[width=\linewidth]{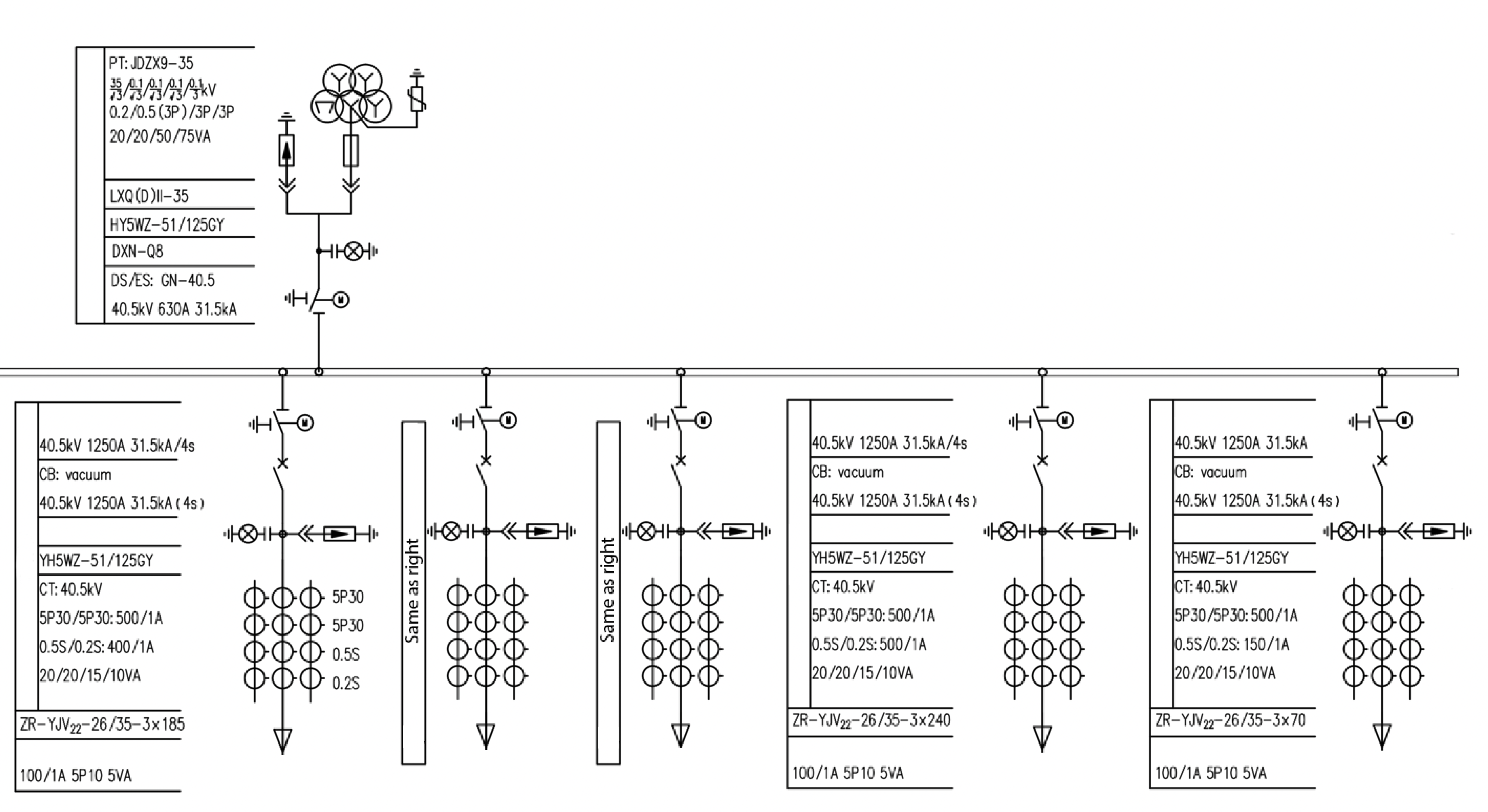}
    \caption{An example of electrical engineering drawing. There are 6 circuits and 6 tables in this electrical engineering drawing. 
    The engineering drawing has been modified to prevent copyright infringements.}
    \label{fig:drawing_demo}
\end{figure*}

In general, a single electrical engineering drawing includes multiple circuits and tables as depicted in Fig.~\ref{fig:drawing_demo}. Each circuit describes different electrical devices, while each table presents the parameter settings for the corresponding circuits. 
To enhance the identification of circuits and tables, it is necessary to extract these components from diagrams and analyze them using specialized models. While there is a considerable amount of existing work on detecting and classifying specific objects within images \cite{lin2017feature,ultralytics2020yolov5,carion2020end}, the correspondence between these objects is often lost in this process. 
Most current methods for extracting relations within documents depend on the relations between text pairs \cite{xu2020layoutlm,yu2021pick,peng2022ernie}, rendering them insufficient for addressing relations between diagrams and tables. 
In visual relation detection methods \cite{liang2018visual,abdelkarim2021exploring,zhao2023unified}, the task and model design limit their capability to predict relationships across all circuit-table pairs, thereby causing missed detections.

The circuit-to-table relations in electrical engineering drawings are complex, potentially exhibiting one-to-one, one-to-many, and other variations. 
Furthermore, the quantity of circuits and tables in each image differs, presenting difficulties in managing variable-length inputs using non-sequential modeling methods. 
To address these issues, we propose \name, a \textbf{Vi}sual \textbf{R}elation prediction model for \textbf{E}ngineering \textbf{D}rawings based on Transformer architecture \cite{vaswani2017attention}. 
\name consists of a visual encoder, an object encoder, and a relationship decoder. 
The model is trained and fine-tuned utilizing the PubLayNet \cite{zhong2019publaynet} dataset along with the proprietary electrical engineering drawing dataset, resulting in exceptional performance in the task of relation prediction.

The main contributions of this work are outlined as follows:
\begin{itemize}
    \item We present a novel vision-based relation detection approach, named \name, to address the issue of predicting relations for non-textual components in complex documents. This approach has been specifically implemented for the purpose of circuit-to-table relation matching in electrical design drawings.
    \item We develop a dataset of electrical engineering drawings derived from industrial design data, and we annotate the instances and their relationships within the dataset.
    \item We evaluate our method using various metrics on the electrical engineering drawing dataset. Furthermore, we perform comparative analyses with existing approaches and provide a performance comparison between the existing methods and our proposed technique.
    \item We perform extensive ablation studies to compare the impact of different model architectures, hyperparameters, and training methods on the overall performance. Moreover, we refined our model architecture based on these comprehensive and comparative analysis.
\end{itemize}

The following sections of this paper are structured as follows. Section~\ref{sec:related_works} reviews the related work pertinent to this study. Section~\ref{sec:methodology} elaborates on the proposed methods and models in detail. Section~\ref{sec:experiments} discusses the experimental evaluation results of the proposed approach. Lastly, the article ends with a summary of our work in Section~\ref{sec:conclusion}.

\section{Related Works}
\label{sec:related_works}
\subsection{Visual Document Understanding}
The task of visual document understanding (VDU) focuses on understanding digital documents in image formats. There are several downstream tasks associated with VDU, including key information extraction, relation detection, document layout analysis, and others. 
Most contemporary VDU techniques rely on deep neural networks that utilize visual, textual, or a mixture of visual and textual modalities.

Approaches for document understanding that utilize computer vision were initially proposed following advancements in convolutional neural networks (CNN). Hao et al. \cite{hao2016table} proposed the use of CNN for detecting tables in document images. 
With the introduction of the R-CNN series model \cite{girshick2014rich,girshick2015fast,ren2015faster,he2017mask}, the methodologies for detecting tables \cite{schreiber2017deepdesrt,prasad2020cascadetabnet,fernandes2022tabledet} and analyzing layouts \cite{soto2019visual,zhong2019publaynet} in document images have benefited from these models, resulting in enhanced performance.
Modern Optical Character Recognition (OCR) engines \cite{smith2007overview} have demonstrated significant efficacy in extracting text from document images. Furthermore, the advancement of language models has contributed to the tasks of document understanding. However, recent approaches in VDU incorporate other modalities such as textual information as auxiliary information. 
LayoutLM~\cite{xu2020layoutlm} utilizes textual data in conjunction with the bounding box provided by the OCR engine for inference. DocFormer~\cite{appalaraju2021docformer}, LayoutLMv2~\cite{xu2020layoutlmv2} and LayoutLMv3~\cite{huang2022layoutlmv3} utilize early fusion techniques to better integrate and leverage visual and textual modality information.

\begin{figure*}[t]
    \centering
    \includegraphics[width=\linewidth]{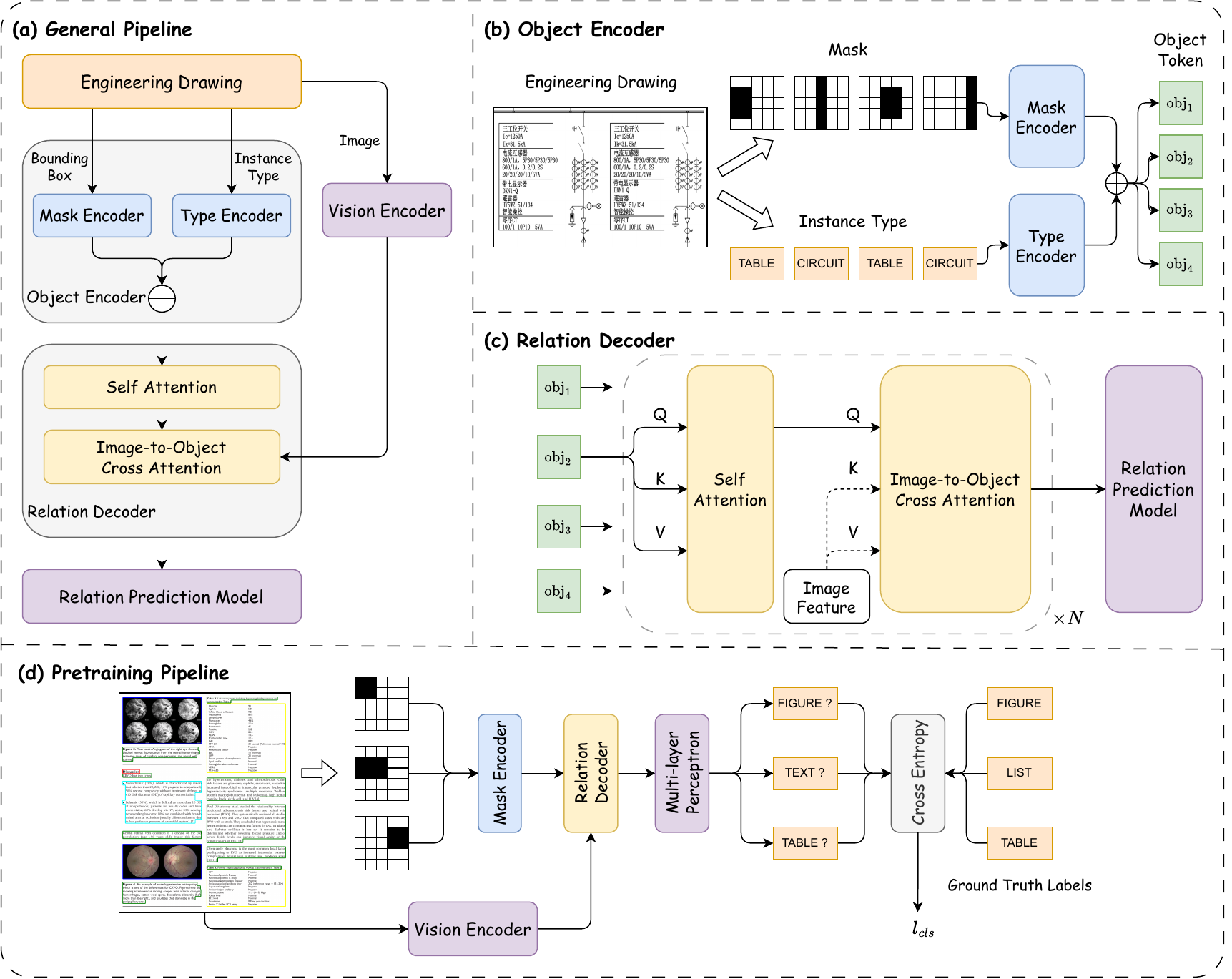}
    \caption{Overview of the general pipeline of \name.
    (a) Engineering drawings are processed through the Vision Encoder, Object Encoder, Relation Decoder, and Relation Prediction Model.
    (b) The Object Encoder converts the instance masks and types into mask and type embeddings, which are then aggregated to form the object tokens.
    (c) The Relation Decoder utilizes the object tokens as inputs and integrates them with the image features from the Vision Encoder through a cross-attention mechanism.
    Residual connections between layers are ignored for simplicity.
    (d) While pretraining, the model encodes the document images and position masks, and after decoding through the relation decoder, it predicts the image classification of the position where the mask is located.
    }
    \label{fig:general_pipeline}
\end{figure*}

\subsection{Visual Relation Detection}
Visual relation detection (VRD) involves the task of predicting the relations or interactions between pairs of objects within a single image~\cite{cheng2022visual}. Typically, VDR is a mid-level vision task that extracts information from low-level vision tasks such as object detection and recognition~\cite{lu2016visual}.
Despite the advancement of VDR techniques, the task remains difficult because of conflicts arising from the various potential relations and the lack of labeled data~\cite{liang2018visual}.

Lu et al.~\cite{lu2016visual} introduced the first VDR approach utilizing deep learning techniques. This method employed R-CNN for object and predicate detection and incorporated language priors to enhance the accuracy of predicted relations.
Subsequently, ViP-CNN was introduced by Li et al.~\cite{li2017vip} to identify subject, predicate, and object simultaneously. This method incorporates a Phrase-Guided Message Passing Structure to investigate the interconnections of relation components.
Zhuang et al.~\cite{zhuang2017towards} proposed a context-aware interaction classification framework based on an attention mechanism, which is accurate, scalable, and enjoys good generalization ability to recognize unseen context-interaction combinations.
Though these methods succeeded in extracting relations in general images, few studies~\cite{li2019engineering} have investigated the use of VRD techniques in domain-specific images, such as structured documents, mechanical blueprints, and engineering drawings.

\section{Methodology}
\label{sec:methodology}
\subsection{Problem Definition}
Given an image of an electrical engineering drawing, let there be $N_c$ circuits and $N_t$ tables present within the image. There may be a relation between a circuit and a table.
That is, there may exist at most $N_c\times N_t$ relations in the engineering drawing. 
Assuming the presence of bounding boxes and their associated instance type labels, we aim to determine if there exists a relation between specific circuits and tables.

\subsection{Model Architecture}

\name has three main components: a pretrained vision encoder, a lightweight object encoder, and a fast relation decoder. 
Fig.~\ref{fig:general_pipeline} presents the overview of our model pipeline.
We will describe these components in the following sections.

\subsubsection{Vision encoder}

The vision encoder in the model processes an image $I_{doc}$ of size $3\times H_{\textrm{Image}}\times W_{\textrm{Image}}$ and produces an image embedding in either feature vector format $F_{\textrm{vision}}$ or feature map format with dimensions $C_{v}\times H_{v}\times W_{v}$, where $C$, $H$, and $W$ represent the channels, width, and height of the image and feature map. In cases where feature maps are produced, they are flattened and converted into feature vectors. To prevent bias, a pre-trained vision encoder called the Masked Autoencoder (MAE)~\cite{he2022masked} is utilized in this study. The Document Image Transformer (DiT)~\cite{li2022dit} is employed as our vision encoder. The vision encoder is only used once for each image, regardless of the number of objects or relationships in the image.

\begin{equation}
\label{eq:vit1}
F_{\textrm{Vision}}=\textrm{Vision-Encoder}(I_{doc})
\end{equation}
\begin{equation}
\label{eq:vit2}
F_{\textrm{Vision}}=W\cdot\textrm{Flatten}(\textrm{Vision-Encoder}(I_{doc}))
\end{equation}

\subsubsection{Object encoder}


The object encoder efficiently maps a bounding box and its typing information into a vectorial embedding. 
We use a convolutional neural network (CNN) to encode the bounding boxes. The object's bounding box is shown as a one-channel image $I_{obj}$ with the same dimensions as the engineering drawing image. Pixels inside the bounding box are shown as 1, while those outside are shown as 0. The bounding box image is then encoded using a three-layer CNN.
\begin{equation}
\label{eq:mask}
F_{\textrm{mask}}=\textrm{CNN}(I_{obj})
\end{equation}

To inform the relation decoder about whether the embedding token represents a circuit or a table, we aggregate the vector embeddings of bounding boxes $F_{\textrm{mask}}$ with two learned embeddings $F_{\textrm{type}}$.

\begin{equation}
\label{eq:obj}
F_{\textrm{object}}=F_{\textrm{mask}}+F_{\textrm{type}}
\end{equation}

\subsubsection{Relation decoder}
The relation decoder takes the encoded masks and vectorized images as input and predicts if there is a relation between each circuit-table pair. In detail, the relation decoder consists of two components: the fusion model and the relation prediction model.

The fusion model is employed for feature fusion between object masks and image features. 
Inspired by DETR~\cite{carion2020end}, a transformer-decoder-based model is adopted. 
We modify the standard transformer decoder by eliminating the relative position embedding and causal mask because of the lack of order between objects. This transforms the decoder into a bidirectional transformer decoder. 
As shown in Fig.~\ref{fig:general_pipeline}, each decoder layer consists of four parts.
\begin{enumerate*}
    \item The tokens are initially processed by a self-attention model as shown in Eq.~\ref{eq:self_attn}, which enables the mask tokens to interact with each other. This step enables the objects to determine their relative positions. 
    \item Next, a cross-attention model is introduced, where the image embeddings are used as the key and value vector, and mask tokens are used as the query vector like Eq.~\ref{eq:cross_attn}. By utilizing the image-to-object cross-attention model, the mask tokens are modified through a combination of bounding box features and vision features. 
    \item Then, a feedforward layer updates all the mask tokens, and a dropout layer is used to improve the model's generalizability. 
    \item Finally, a residual connection is added to every attention layer and feed-forward layer, in accordance with the typical transformer architecture~\cite{vaswani2017attention}.
\end{enumerate*}

\begin{equation}
\label{eq:self_attn}
F_{\textrm{object}}=\textrm{Attention}(F_{\textrm{object}},F_{\textrm{object}},F_{\textrm{object}})
\end{equation}
\begin{equation}
\label{eq:cross_attn}
F_{\textrm{object}}=\textrm{Attention}(F_{\textrm{object}},F_{\textrm{vision}},F_{\textrm{vision}})
\end{equation}

\begin{figure}[t]
    \centering
    \includegraphics[width=\linewidth]{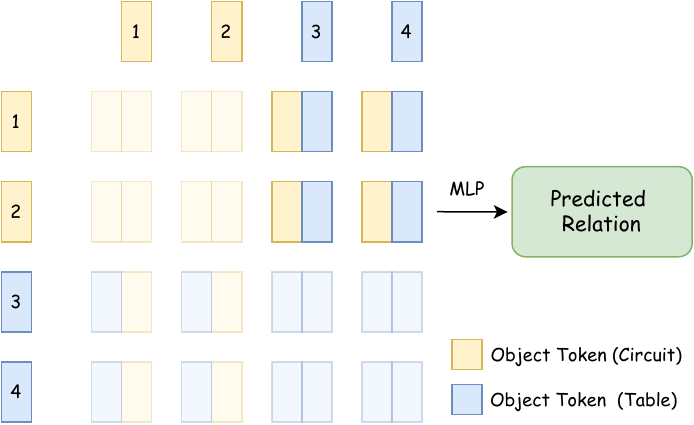}
    \caption{The implementation of relation prediction model. The semi-transparent tokens represent the filtered parts, which do not participate in the relationship prediction computation.
    }
    \label{fig:rel_pred}
\end{figure}

After receiving the object tokens from the fusion model of the relation decoder, it is essential to identify the relationships between them using the relation prediction model. The combined object tokens are concatenated to generate $(N_{c}+N_{t})^2$ combination tokens. 
This process is depicted in Fig.~\ref{fig:rel_pred}. 
To streamline the training and inference procedures, redundant and unfeasible combinations are eliminated. The relation prediction model consists of a three-layer perceptron with ReLU activation and a linear projection layer. This model is utilized to convert hidden features into a two-dimensional logit output. Afterwards, the main objective of the relation prediction model is to determine whether a relationship exists between two objects.

\begin{equation}
\label{eq:rel_concat}
F_{\textrm{relation}_{i,j}}=F_{\textrm{object}_i}\oplus F_{\textrm{object}_j}
\end{equation}
\begin{equation}
\label{eq:rel_pred}
\textrm{Relation-Prediction}=\textrm{MLP}(F_{\textrm{relation}})
\end{equation}

\subsection{Pretraining}

\begin{figure*}[!htb]
\newcommand{\scale}{0.45}
\centering
\subfloat[A.1]
{
\includegraphics[width=\scale\linewidth]{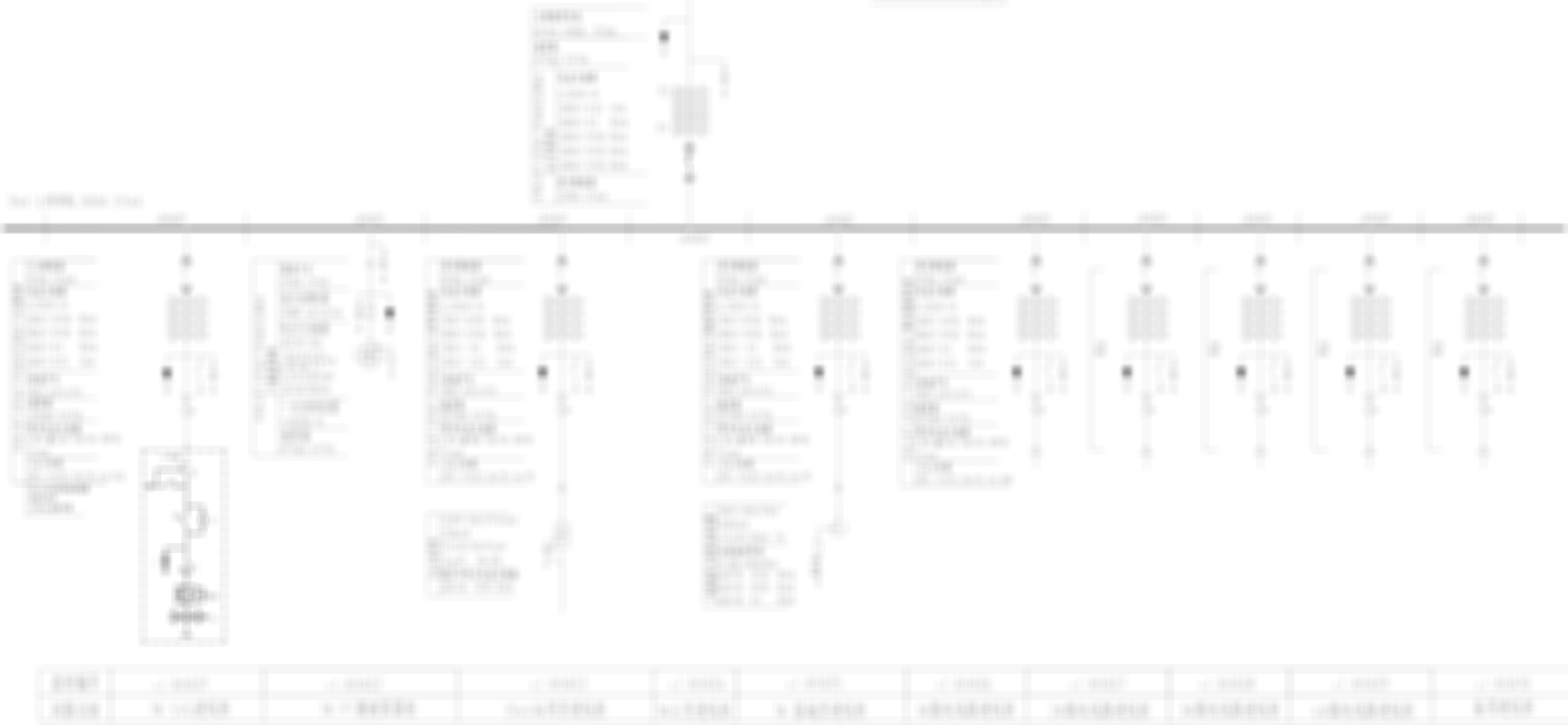}
}
\subfloat[A.2]
{
\includegraphics[width=\scale\linewidth]{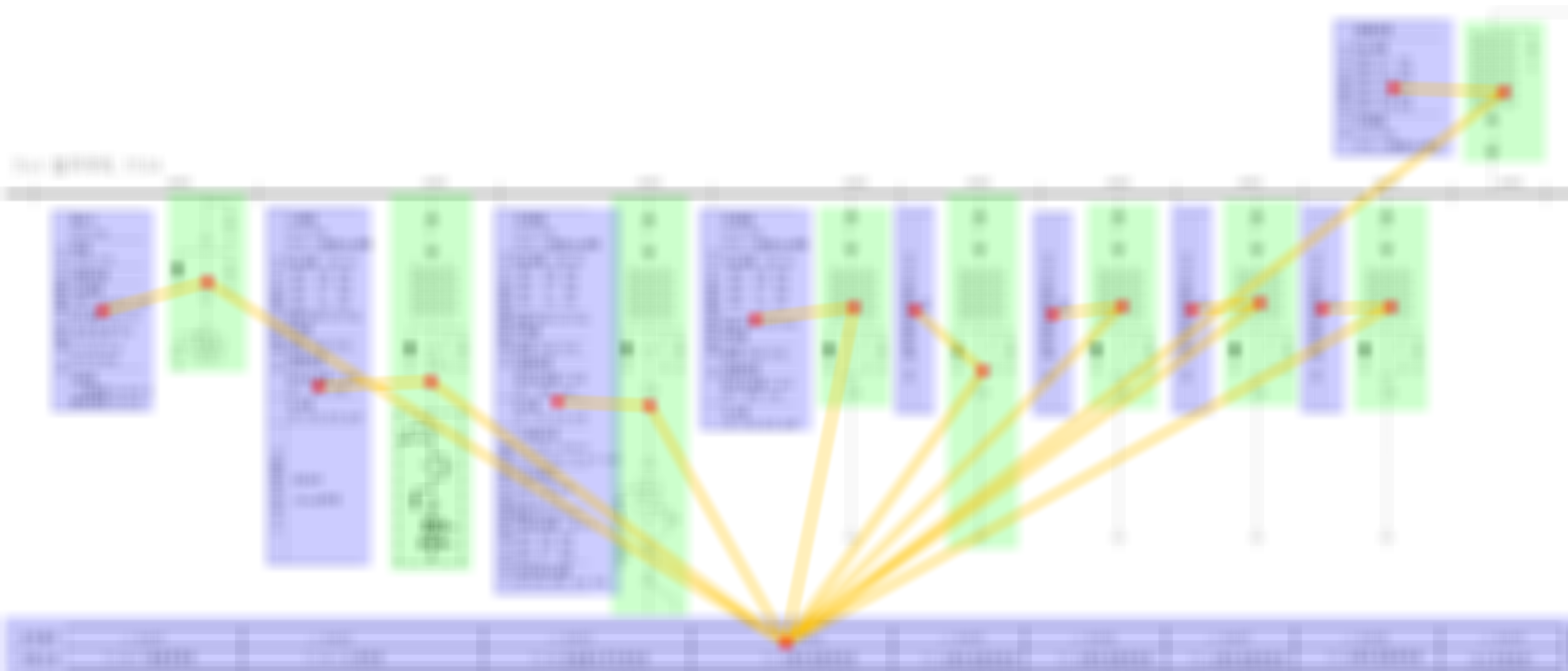}
}
\\
\subfloat[B.1]
{
\includegraphics[width=\scale\linewidth]{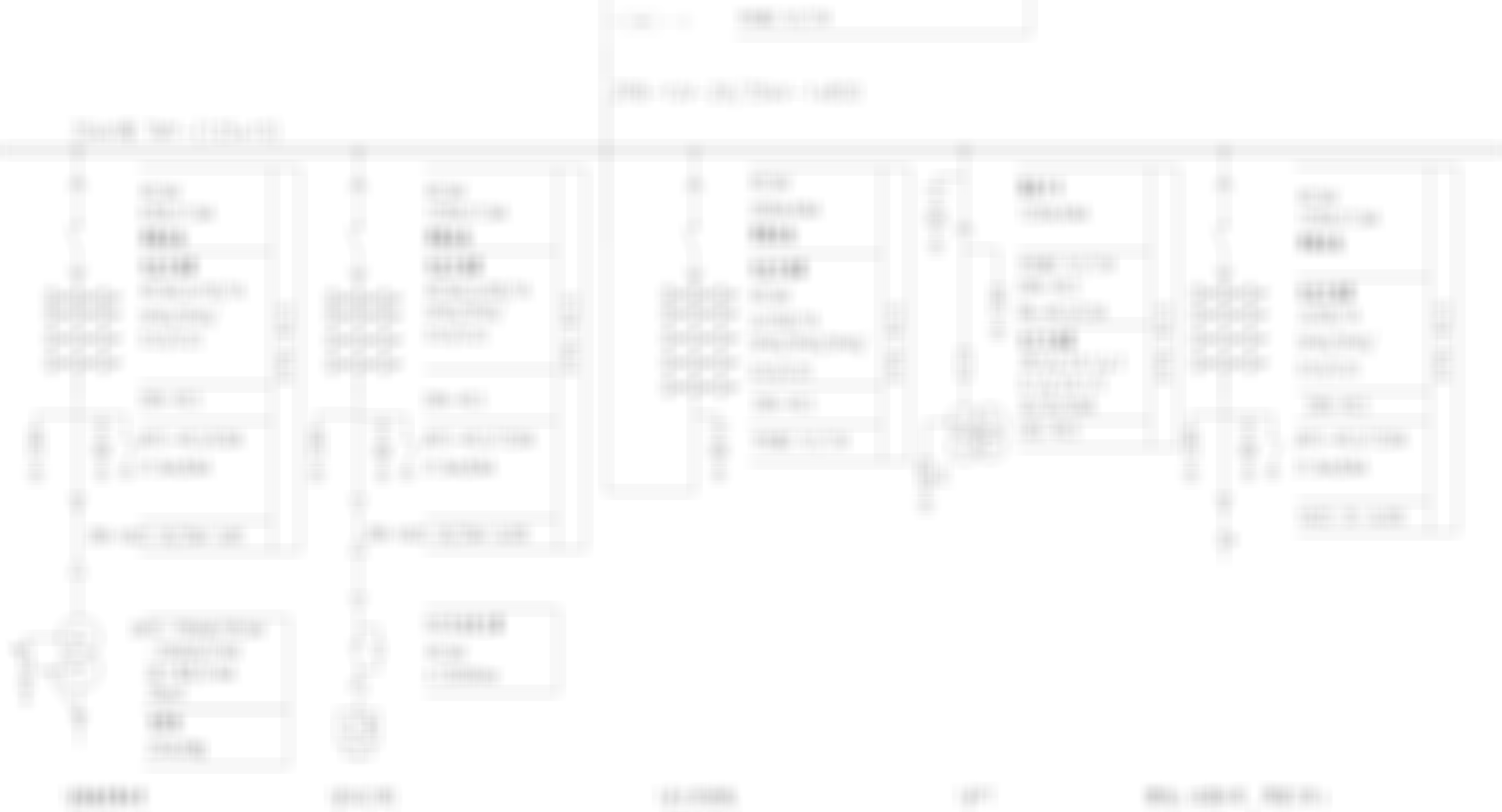}
}
\subfloat[B.2]
{
\includegraphics[width=\scale\linewidth]{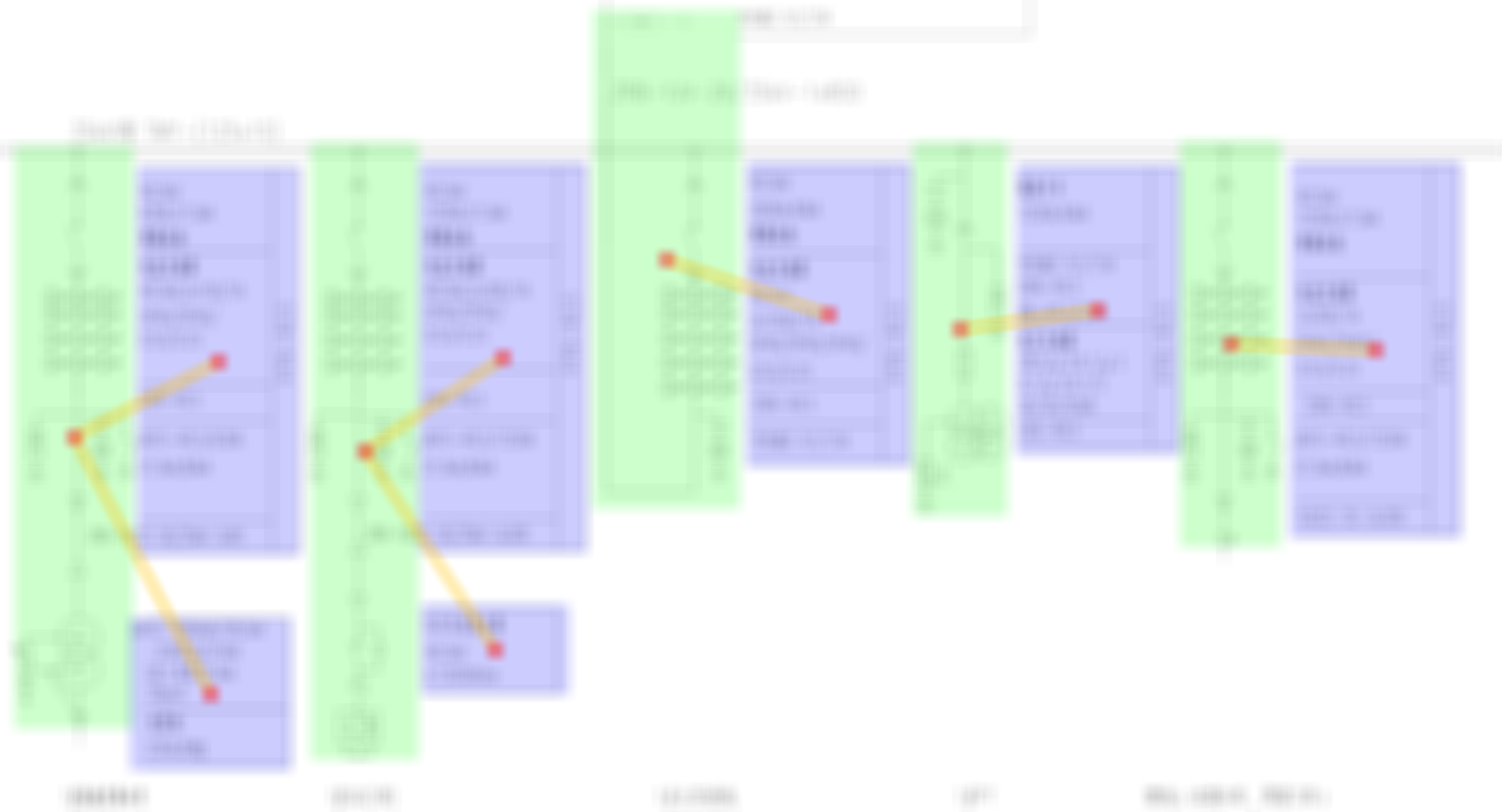}
}
\\
\subfloat[C.1]
{
\includegraphics[width=\scale\linewidth]{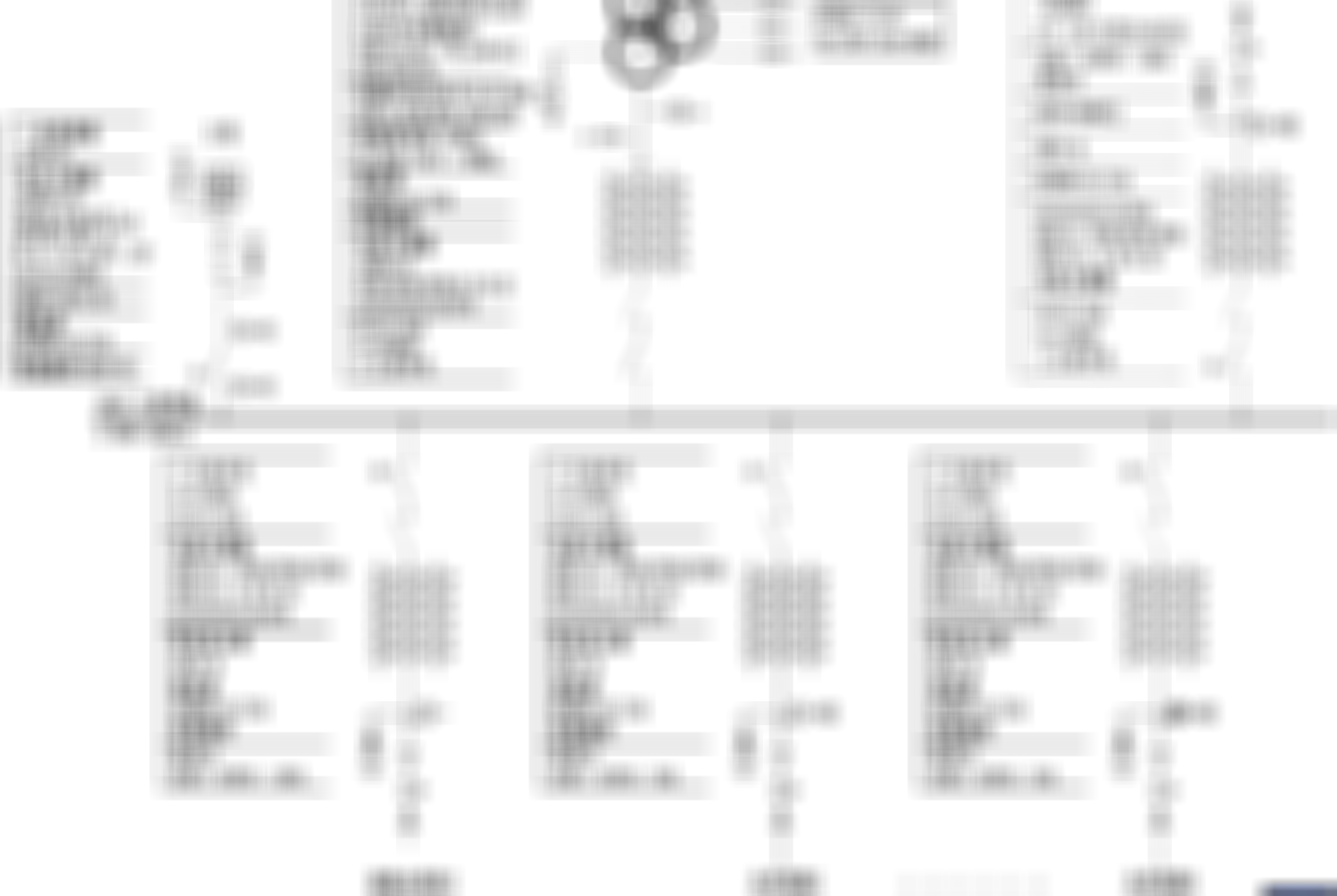}
}
\subfloat[C.2]
{
\includegraphics[width=\scale\linewidth]{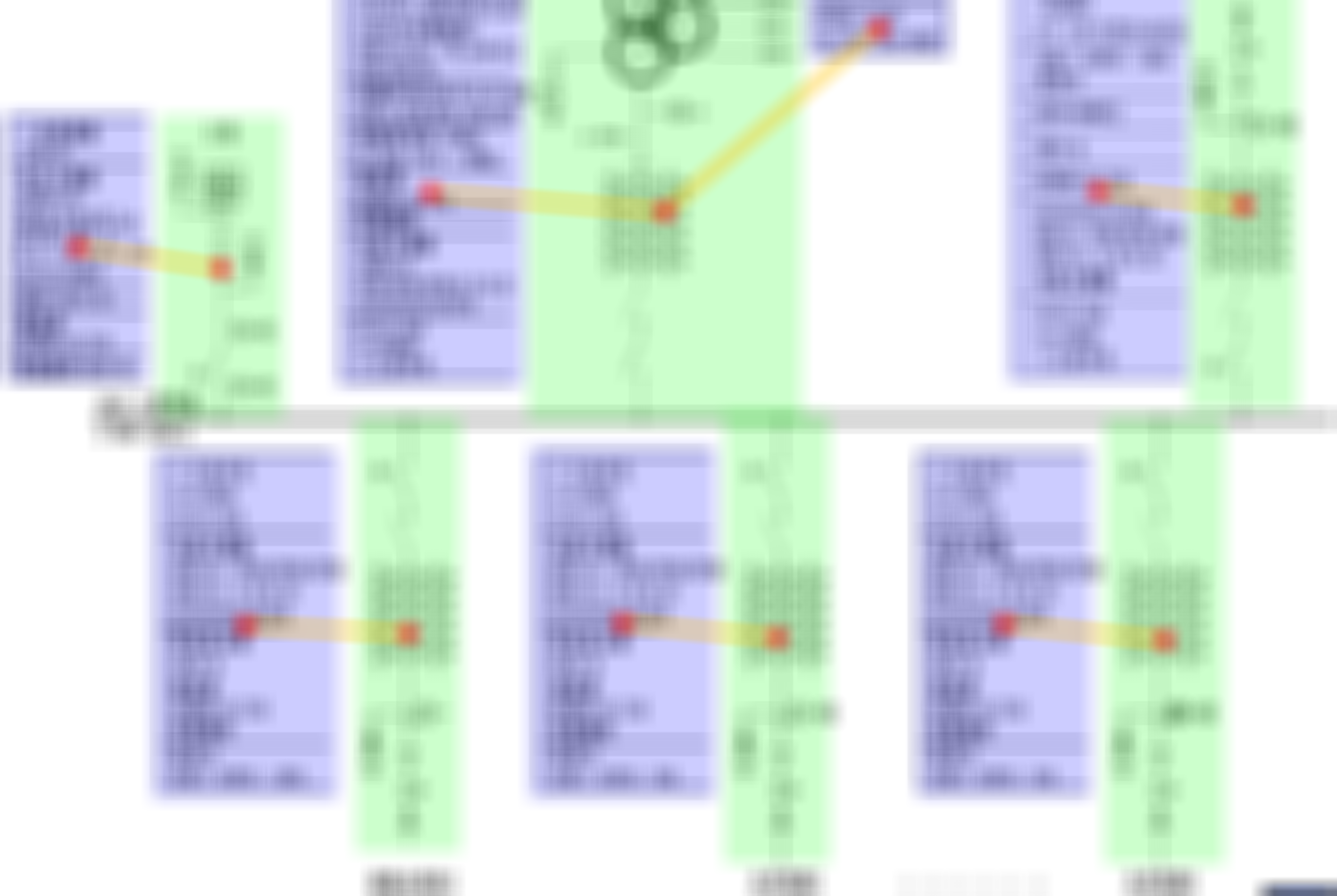}
}
\\
\subfloat[D.1]
{
\includegraphics[width=\scale\linewidth]{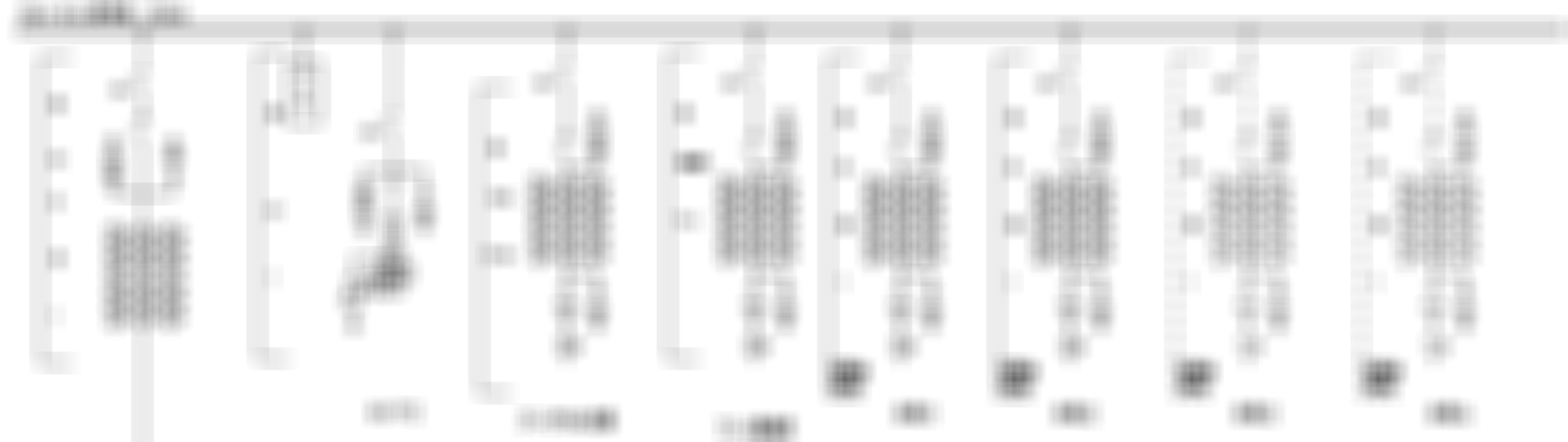}
}
\subfloat[D.2]
{
\includegraphics[width=\scale\linewidth]{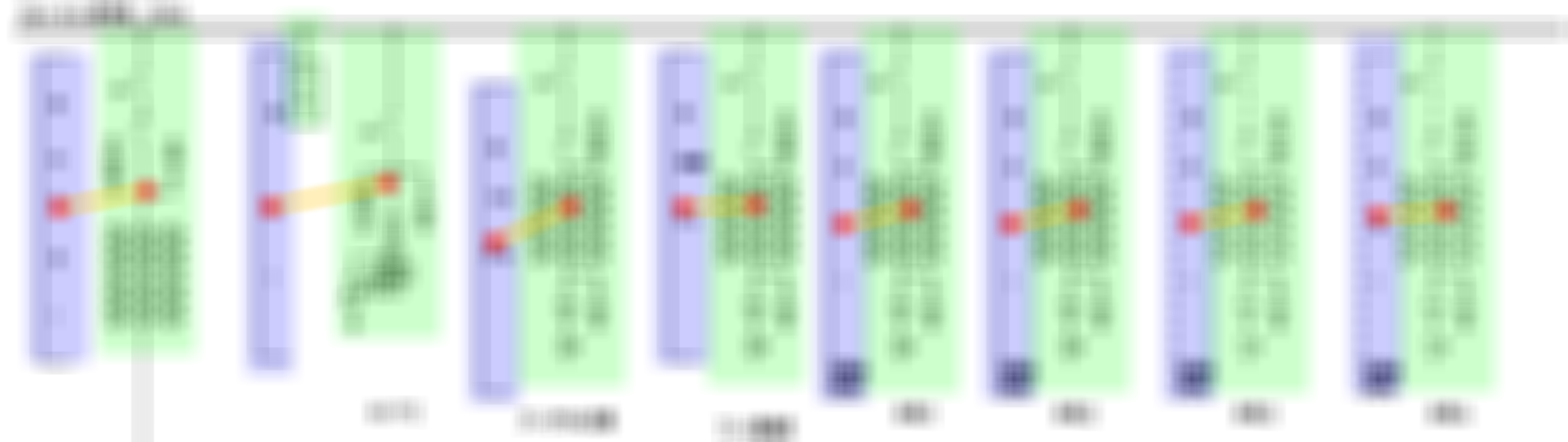}
}

\caption{Qualitative result of our model. Caption X.1 and X.2 denotes the original electrical engineering drawings and the model prediction results, respectively.}
\label{fig:qualitative}
\end{figure*}

Due to the scarcity of annotated training data for relation detection in engineering drawings, we utilize the PubLayNet~\cite{zhong2019publaynet} dataset during the pretraining phase. 
PubLayNet offers a dataset consisting of 340,000 document images, including 3 million bounding boxes and type annotations for the instances within these images. 
To enhance the model's comprehension of positional data, we devise a pretraining task that employs our model to categorize the instance type of the masked region as shown in Fig.~\ref{fig:general_pipeline}-(d).
The architecture of our pretraining model exhibits minor variations compared to the relation prediction model.
Since it is necessary to predict the category of the provided bounding boxes (i.e., table or circuit), the type embeddings integrated into the vector embeddings of bounding boxes within the object encoder are excluded. 
The relation prediction model is substituted with a multi-layer perceptron to perform classification tasks.


\section{Experiments}
\label{sec:experiments}

\begin{table}[ht]
\caption{
Comparison with visual-base relation prediction methods on our validation dataset.
}
\centering
\scalebox{1.2}{
\begin{tabular}{cc|c}
\toprule
\textbf{Method} & \textbf{Architecture} & Accuracy$(\%,\uparrow)$ \\
\midrule
RVL-BERT\cite{chiou2021visual} & Transformer & 46.7 \\
VTransE\cite{zhang2017visual} & TransE + CNN & 87.8 \\
MBBR\cite{anastasakis2024self} & Transformer & 86.8  \\
\name (\textbf{Ours}) & Transformer & \textbf{96.3}  \\
\bottomrule
\end{tabular}
}
\label{tab:comparison}
\end{table}

\subsection{Experiment Setup}
\subsubsection{Dataset}
In the pretraining phase, our method is trained and tested on the PubLayNet datasets with the same train and test splits as the original dataset. 
For the finetuning phase, we used a self-annotated dataset of engineering design diagrams containing 283 images, 4,566 entities, and 2,112 relations between entities.
The datasets are randomly divided, with 90\% assigned for training and 10\% for evaluation.

\subsubsection{Implementation Details}
The dimension $D$ of latent representation throughout the model pipeline is set to 768. 
For the vision encoder backbone, we adopt the DINOv2-B model \cite{oquab2023dinov2}. 
The model consists of 12 layers of Transformer encoders, each containing a 12-head multi-head attention block. The input image is cropped into patches of size 14, which are then processed and encoded to a 768-dimensional feature vector.
We finetuned the backbone network using the same MAE training method as DiT \cite{li2022dit} based on its state-of-the-art performance on visual-based document downstream tasks.
The bounding box encoder is a three-layer convolution network with a ReLU activation function and a linear projection layer that maps the feature map to a 768-dimensional vectorized feature.
The relation decoder is a 2-layer Transformer Decoder for multi-modal feature fusion.
The input image size for the vision encoder is $518\times518$, while the object encoder takes $224\times224$ masks as input. Despite the difference in input sizes, we maintain the relative positions between the images and the masks to help the model better understand the positional relations between them.

We implement our model in PyTorch.
For the pretraining phase, our model is trained with a batch size of 64 for a total of 100 epochs on the PubLayNet dataset using $\sim$90 NVIDIA A800 GPU hours. 
During the finetuning phase, we initialized the type embeddings to zero and the multi-layer perceptron of the relation extractor with uniformly random numbers. Meanwhile, other parameters remained unchanged from the pretraining checkpoint. 

We applied data augmentation to the data while training, including horizontal and vertical flipping with a probability of 0.2, $D_4$ dihedral group transformations, and fixed-size random cropping. Unlike common random cropping, we ensured that instances in the images were not lost due to cropping.

For optimization, we use AdamW with hyper-parameters $\beta=(0.9,0.999),\epsilon=10^{-8}$ and weight decay of $10^{-3}$. The base learning rate for the optimizer is $10^{-4}$, we employ a linear learning rate schedule from $10^{-4}$ to $10^{-5}$ with 20\% epoches for warming up.

\subsection{Metrics}
Unlike the commonly used R@x metric in relation detection tasks \cite{lu2016visual}, we use mAP, precision, and recall to assess model performance in relation detection tasks. Standard relation detection tasks may not annotate all possible relations in an image. In our task, relations between tables and circuits are clear, so using mAP is a more accurate indicator of model performance. Accuracy is used as the metric for comparison experiments.

\subsection{Experiment Results}
\subsubsection{Qualitative Result}
Qualitative results are shown in Fig.~\ref{fig:qualitative}. The blue and green areas represent the bounding box regions of circuits and tables, respectively. In the presence of a relationship between a table and a circuit, a line segment will link the two entities.

\subsubsection{Comparison}
In this section, we evaluate our model against current methods for relation detection. We focus on visual relation detection instead of text-based methods for document relation extraction. To align with the visual task, we modify the dataset by categorizing textual descriptions as either "table" or "circuit." Relations are established between all circuit-table pairs, with existing relations labeled as "describe" and non-existing relations labeled as "not describe."

Table~\ref{tab:comparison} provides a comparative analysis of our relation prediction method in contrast to several established approaches, with the accuracy metric used to measure the precision of the outcomes. Our findings demonstrate that our method surpasses the results of our engineering drawing relation prediction dataset. Importantly, we employ the evaluation methods used by the compared approaches to maintain consistency with our metrics. For MBBR, the accuracy is calculated based on the top-100 relation predictions.


\begin{table*}[ht]
\caption{
Ablation study of model architecture. The precision and recall metrics report data only for the "has relation" category, while mAP provides the overall average value.
}
\centering
\scalebox{1.17}{
\begin{tabular}{cc|cc|ccc}
\toprule
\textbf{Vision Encoder} & \textbf{Object Encoder} & \textbf{Image Size} & \textbf{\#Rel. Decoder Layer} &  mAP$(\%,\uparrow)$ & Precision$(\%,\uparrow)$ & Recall$(\%,\uparrow)$\\
\midrule
ViT-B/14 & 3-Layer CNN & $518^2$ & 2 & 89.00 & \textbf{88.41} & 76.39 \\
ViT-S/14 & 3-Layer CNN & $518^2$ & 2 & 86.37 & 85.83 & 59.27 \\
ViT-L/14 & 3-Layer CNN & $518^2$ & 2 & \textbf{89.80} & 82.79 & 72.66 \\
ResNet-50 & 3-Layer CNN & $518^2$ & 2 & 84.33 & 75.94 & 68.47 \\
\midrule
ViT-B/14 & 3-Layer CNN & $518^2$ & 1 & 84.77 & 81.20 & 75.05 \\
ViT-B/14 & 3-Layer CNN & $518^2$ & 3 & 88.91 & 85.37 & 79.56 \\
\midrule
ViT-B/14 & 6-Layer ViT & $518^2$ & 2 & 59.57 & 50.00 & 14.07 \\
\midrule
ViT-B/14 & 3-Layer CNN & $294^2$ & 2 & 88.16 & 85.03 & 79.18 \\
ViT-B/14 & 3-Layer CNN & $756^2$ & 2 & 87.52 & 86.55 & \textbf{80.91} \\
\bottomrule
\end{tabular}
}

\label{tab:ablation_arch}
\end{table*}

\subsubsection{Ablation study}
In this section, we evaluate our method in different settings and strategies. 

We conduct an additive ablation study on the mAP, precision, and recall for relation prediction on the validation set, as illustrated in Tab.~\ref{tab:ablation}. 
In the baseline model, we initialize the model with random parameters without pretraining and omit the type embedding from the object encoder. Introducing type embedding allows the model to better differentiate the categories of object tokens. When computing self-attention over object tokens, the relation decoder can simultaneously consider the categories of the objects, thereby reducing computations for improbable relations and enhancing the understanding of relations. Due to the insufficient data in our task’s dataset, we introduced a pretraining mechanism for both the vision encoder and the position encoder-decoder.
By undergoing unsupervised training on a large-scale document image dataset, the vision encoder more effectively extracts visual features from electrical engineering drawings. Supervised pretraining was utilized to predict the classification of selected regions based on given positional information, thereby enhancing the model's understanding of positional inputs. As shown in Table~\ref{tab:ablation}, the inclusion of type embedding and model pretraining significantly improved the performance across all evaluation metrics.

\begin{table}[ht]
\caption{
Additive ablation study of sequentially applying different training techniques for relation prediction on valid split of electrical engineering dataset
}
\centering
\begin{tabular}{l|ccc}
\toprule
\textbf{Ablate} & mAP$(\%,\uparrow)$ & Precision$(\%,\uparrow)$ & Recall$(\%,\uparrow)$ \\
\midrule
Base model & 88.26 & 81.52 & 75.44 \\
+\xspace Type Embedding & 90.48 & 81.15 & 76.27 \\
+\xspace DiT-style Pretrain & 88.93 & 85.69 & 79.16 \\
+\xspace Position Pretrain & 91.65 & 90.11 & 83.23 \\
\bottomrule
\end{tabular}
\label{tab:ablation}
\end{table}

 We also conducted ablation experiments on the model architecture. We compared the effects of various vision encoders, diverse object encoders, different input image resolutions, and different numbers of transformer decoder layers in the relation decoder on the performance.
 The experimental results are shown in Tab.~\ref{tab:ablation_arch}. The hidden dimension of the model is determined by the output dimension of the vision encoder. For ViT-S, ViT-B, ViT-L \cite{oquab2023dinov2}, and ResNet-50~\cite{he2016deep}, the dimensions for latent representation are set to 384, 768, 1024, and 768, respectively.
Upon analyzing the experimental results, it becomes evident that the object encoder architecture significantly influences model performance. The relation decoder struggles to interpret object tokens encoded by ViT, resulting in a notable decline in performance across various metrics. 
Additionally, the vision encoder impacts the latent representation dimension of the model, thereby also playing a crucial role in determining model performance. The findings suggest that ViT-S performs poorly due to its constrained embedding dimension. For ViT-B and ViT-L, the limited task difficulty results in negligible performance differences. 
Higher image resolutions improve the vision encoder's capacity to capture more detailed image features. However, the vision encoder’s ability to encode all these details is restricted by the dimension of the image feature vector, leading to a minimal effect on performance.

\subsubsection{Inference Efficiency}
\begin{figure}[t]
    \centering
    \includegraphics[width=\linewidth]{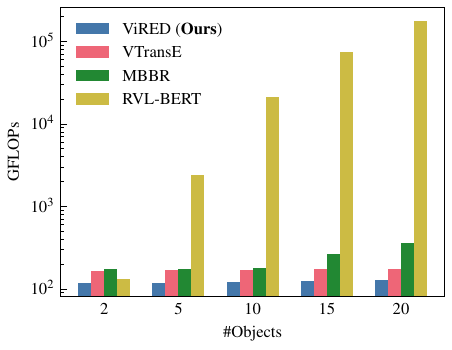}
    \caption{FLOPs of different models with respect to different numbers of objects.}
    \label{fig:flops}
\end{figure}

To evaluate the inference efficiency of the model, we compute the FLOPs (Floating Point Operations) for the inference process. For uniformity, a batch size of 1 is employed for all models. The input image resolution is fixed at $518\times518$, with the number of objects to be predicted ($N$) ranging from 1 to 20. The relations to be predicted include all possible interactions between objects, amounting to $(N-1)^2$. The results are presented in Fig.~\ref{fig:flops}.
Owing to the utilization of a lightweight object encoder and relation decoder, the inference efficiency of \name is minimally impacted by the number of objects in the electrical engineering drawing. It sustains a rapid inference speed even when a single drawing contains numerous objects.

\section{Conclusion}
\label{sec:conclusion}
We propose a novel relation prediction method and present a dataset for relation detection in electrical engineering drawings. We apply this relation prediction method to the task of relation detection within the dataset, achieving high performance on the validation set through the processes of pretraining and finetuning the model. 
We conduct a series of experiments, including comparative analyses with existing methods and ablation studies on training strategies and model architectures. The experimental results indicate that our method achieves superior performance on this task.

\newpage

\bibliographystyle{IEEEtran}
\bibliography{IEEEabrv,ref}


\end{document}